\documentclass[runningheads]{llncs}

\usepackage{eccv}

\usepackage{eccvabbrv}

\usepackage{graphicx}
\usepackage{booktabs}

\usepackage[accsupp]{axessibility}  

\usepackage[pagebackref,breaklinks,colorlinks,citecolor=eccvblue]{hyperref}

\usepackage{orcidlink}
\usepackage{multirow}
\usepackage{color, colortbl}
\usepackage{cancel}
\usepackage{array}
\usepackage{hhline}
\newcolumntype{L}[1]{>{\centering\arraybackslash}l{#1}}
\definecolor{LightCyan}{rgb}{0.88,1,1}

\usepackage{xr}
\makeatletter

\newcommand*{\addFileDependency}[1]{
\typeout{(#1)}
\@addtofilelist{#1}

\IfFileExists{#1}{}{\typeout{No file #1.}}
}\makeatother

\newcommand*{\myexternaldocument}[1]{%
\externaldocument{#1}%
\addFileDependency{#1.tex}%
\addFileDependency{#1.aux}%
}

\myexternaldocument{supplementary}

\begin{document}

\title{EasyPortrait -- Face Parsing and Portrait Segmentation Dataset} 

\titlerunning{}


\author{Karina Kvanchiani \and 
Elizaveta Petrova\orcidlink{0009-0005-8454-7701
} \and
Karen Efremyan\orcidlink{0009-0007-1606-6716} \and
Alexander Sautin \and
Alexander Kapitanov\orcidlink{0009-0009-6359-2019}}

\authorrunning{K.~Kvanchiani et al.}

\institute{SberDevices}

\maketitle

\begin{abstract}
  Recently, video conferencing apps have become functional by accomplishing such computer vision-based features as real-time background removal and face beautification. Limited variability in existing portrait segmentation and face parsing datasets, including head poses, ethnicity, scenes, and occlusions specific to video conferencing, motivated us to create a new dataset, EasyPortrait, for these tasks simultaneously. It contains 40,000 primarily indoor photos repeating video meeting scenarios with 13,705 unique users and fine-grained segmentation masks separated into 9 classes. Inappropriate annotation masks from other datasets caused a revision of annotator guidelines, resulting in EasyPortrait's ability to process cases, such as teeth whitening and skin smoothing. The pipeline for data mining and high-quality mask annotation via crowdsourcing is also proposed in this paper. In the ablation study experiments, we proved the importance of data quantity and diversity in head poses in our dataset for the effective learning of the model. The cross-dataset evaluation experiments confirmed the best domain generalization ability among portrait segmentation datasets. Moreover, we demonstrate the simplicity of training segmentation models on EasyPortrait without extra training tricks. The proposed dataset and trained models are publicly available\footnote{\url{https://github.com/hukenovs/easyportrait}}.
\end{abstract}

\section{Introduction}
\label{sec:intro}

\begin{figure}
  \centering
  \includegraphics[width=.75\linewidth]{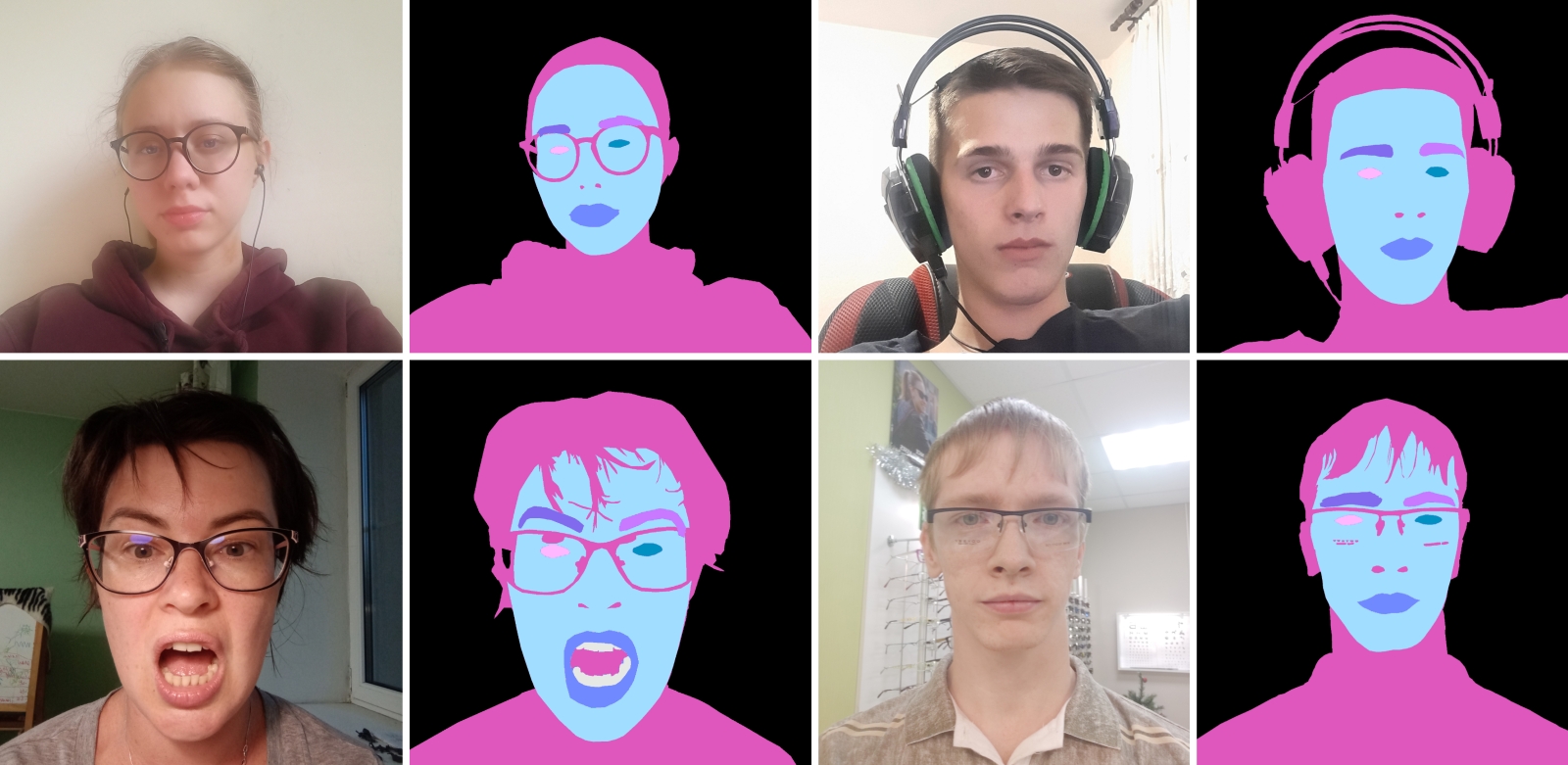}
  \caption{The face parsing and portrait segmentation annotation examples from the EasyPortrait dataset.}
  \label{fig: main}
\end{figure}

Video conferencing apps have quickly gained popularity in corporations for online meetings \cite{covid} and in our habitual life to keep in touch with relatives from a distance \cite{forbes}. As a result, these apps have been enhanced with various beneficial features, including face beautification, background blur, and noise reduction~\cite{banuba}. Such extensions can improve user experience, allowing them to change the background or smooth their skin~\cite{insider}.

Our work primarily aimed to integrate the described features into the video conferencing app. The app should ensure a real-time CPU-based experience on the user's device and produce a highly accurate response. Besides, our system needs to be robust to the amount of context in images, backgrounds, persons in the frame, their poses, attributes (e.g. race and age), and accessories (headphones, hats, etc.). Finally, it is preferable to improve the users' experience with such functions as teeth whitening and accurate background changing in the case of transparent glasses lenses and uneven hair. 

All these requirements impose restrictions on the solution and training dataset. We decided to exploit deep learning models solving portrait segmentation and face parsing tasks for our purpose. The system must function in real-time without any delays and produce fine-grained segmentation masks. The suitable data is required to be 1) heterogeneous in subjects, their head turns, subject-to-camera distances, scenes, and specific for videoconferencing domain subjects' accessories like eyeglasses and headphones; 2) annotated with main face parsing classes (``skin'', ``brows'', ``eyes'', and ``lips'') and an extra class ``teeth''; 3) accurately annotated for both tasks simultaneously. The last can allow training only one model for all possible use cases and save limited resources for model inference.

Existing datasets are unsuitable for our production purposes due to the limitations described in \cref{sec: related}. It motivated us to create the EasyPortrait dataset -- our first contribution. The second is a pipeline for gathering and labeling images with fine-grained segmentation masks utilizing two crowdsourcing platforms, Yandex.Toloka\footnote{\url{https://toloka.yandex.ru}} and ABC Elementary\footnote{\url{https://elementary.activebc.ru}}. It allowed us to collect 40,000 pairs of images and segmentation masks from 13,705 individuals in domain-suitable scenes with different head poses and various specific videoconferencing app accessories. We considered the importance of ethnic diversity to solve problems based on persons and their facial attributes and collected images from users of various countries and continents. Images were annotated manually by 9 classes according to specially designed rules, which allowed us to cover all described cases.

We checked that such data characteristics as the number of samples and their diversity in head poses have a positive impact on the model’s robustness and effectiveness (\cref{ablation}). The generalization ability of the training data was also assessed through cross-dataset evaluation experiments (\cref{experiments}).

\section{Related Work}
\label{sec: related}
In this section, we discuss existing portrait segmentation and face parsing datasets separately. We will consider them from two points of view: 1) the applicability to the videoconferencing domain and 2) data creation techniques for each segmentation task and their consequences.

\subsection{Portrait Segmentation and Matting Datasets}
\label{subsec: datasetps}
The portrait segmentation task involves labeling every pixel in an image as either ``person'' or ``background''. Since matting annotations can be reduced to binary ones, image matting datasets will also be considered. As videoconferencing apps always take portraits for input and not a person entirely, only datasets with waist-deep people are reviewed. Therefore, other popular image person segmentation and image matting datasets, e.g., P3M-10k~\cite{P3M}, PPR10K~\cite{PPR10K}, PhotoMatte13K/85~\cite{photomatte13k}, Persons Labeled~\cite{persons_labeled}, are not described in this paper. We selected EG1800~\cite{eg1800}, AiSeg~\cite{aiseg}, FVS~\cite{ConferenceVideoSegmentationDataset}, Face Synthetics~\cite{synthetic}, and HumanSeg14K~\cite{paddleseg} as the most widespread and predominantly containing photos of people from the waist up. \cref{tabl:related_datasets} provides the numerical analysis of reviewed datasets.

\begin{table}[tb]
  \caption{Comparison of portrait segmentation and face parsing datasets. Because of the specifics of the tasks, we indicated the task name for the first ones and the number of classes for the second. Several datasets include images of different sizes, so the standard label was provided in the resolution column. 96\% of images in the EasyPortrait are FullHD; see more information in \cref{characteristics}. We also included notes about the annotation method, which can be important regarding label quality. For more transparency, 20 classes in the FaceOcc dataset contain 19 classes from CelebAMask-HQ.
  }
  \label{tab:headings}
  \centering
  \scalebox{0.85}{
  \begin{tabular}{|l|c|c|c|c|c|}
    \hline
    Dataset & Samples & Task / Classes & Resolution & Annotation Method\\
    \hline
    EG1800, 2016~\cite{eg1800} & 1,800 & segmentation & 600 × 800 & Photoshop\\
    FVS, 2018~\cite{ConferenceVideoSegmentationDataset} & 3,600 & segmentation & 640 × 360 & chroma-key\\
    AISeg, 2018~\cite{aiseg} & 34,427 & matting & 600 × 800 & automatically\\
    PP-HumanSeg14K, 2021~\cite{paddleseg} & 14,117 & segmentation & 1280 × 720 & manually\\
    The Face Synthetics, 2021~\cite{synthetic} & 100,000 & segmentation & 512 × 512 & automatically\\
    \hline
    Helen, 2012~\cite{helen} & 2,330 & 11 & 400 × 400 & automatically\\
    LFW-PL, 2013~\cite{lfwpl} & 2,927 & 3 & 250 × 250 & automatically \& refined\\
    CelebAMask-HQ, 2019~\cite{celeba} & 30,000 & 19 & 512 × 512 & manually \& checked \& refined\\
    LaPa, 2020~\cite{lapa} & 22,176 & 11 & LR & automatically \& refined\\
    iBugMask, 2021~\cite{ibugmask} & 22,866 & 11 & HR & manually\\
    The Face Synthetics, 2021~\cite{synthetic} & 100,000 & 19 & 512 × 512 & automatically\\
    FaceOcc, 2022~\cite{faceocc} & 30,000 & 20 & 512 × 512 & manually\\
    \hline
    \rowcolor{LightCyan}
    EasyPortrait, 2023 & 40,000 & 9 & FullHD & manually \& checked\\
  \hline
  \end{tabular}}
  \label{tabl:related_datasets}
\end{table}

\textbf{Content.} Chosen datasets can be divided into three groups in terms of image source: 1) downloaded from websites, 2) collected manually, and 3) generated. The last two allow the data authors to directly determine content on their own. Images of EG1800 and AiSeg were collected from services like Flickr, which made their scenes multi-domain. The Face Synthetics dataset was completely generated, entailing primarily blurred backgrounds, and thus far from in the wild. The manually collected FVS~\cite{ConferenceVideoSegmentationDataset} and PP-HumanSeg14~\cite{paddleseg} portrait segmentation datasets are single-domain with a bias towards videoconferencing. The FVS dataset provides composite images from 10 conference-style green-screen videos and virtual backgrounds. As a result, FVS samples suffer from the remaining green screen around a person in the frame. The PP-HumanSeg14 dataset includes 23 different conference backgrounds and samples of participants performing actions such as waving hands and drinking water. The provided samples contain one or more labeled people with faces blurred for privacy.

\textbf{Annotation.} Since segmentation mask annotation is one of the most challenging problems in the computer vision field, data authors prefer automatic methods. All reviewed datasets except PP-HumanSeg14K were annotated automatically or using Photoshop (see \cref{fig: ps} in the
supplementary material for visual examples). Such methods frequently produce coarse masks that prevent accurate high-frequency parts segmentation (e.g., hair) -- one of the main hardships of background removal in video conferencing. 

\subsection{Face Parsing Datasets}
\label{subsec: datasetfp}

The face parsing task aims to assign pixel-level semantic labels for facial images. Generally, face parsing refers to classifying image pixels, such as brows, eyes, nose, lips, mouth, and skin. We considered several widespread face parsing datasets for our purposes (\cref{tabl:related_datasets} and \cref{fig: fp} in the supplementary material). 

\textbf{Content.} The main limitation of existing face parsing datasets' content is low diversity in head poses and the absence of specific occlusions. Such limitation is obtained by utilizing other datasets or websites like Flickr, as Helen's~\cite{helen} authors performed. They searched for ``portrait'' in various languages to avoid cultural bias, and manually filtered out low-quality and false positive images. The CelebAMask-HQ dataset~\cite{celeba} mainly contains front-facing images of celebrities from Celeba~\cite{celeba_main} with centered heads. Besides, the faces are usually occupy a significant part of the image, thus background information is mainly discarded. The LaPa dataset~\cite{lapa} was designed based on images from the 300W-LP~\cite{zhu2017face} and Megaface~\cite{taherkhani2018deep} datasets. Received faces were aligned and mostly cropped with limited background information preserved. This image collection method was also utilized to create the iBugMask dataset~\cite{ibugmask} containing samples from Helen's training set. The iBugMask authors focused on head pose diversity and augmented images with a synthetic rotation algorithm from 3DFFA~\cite{3ddfa_cleardusk}, which led to huge artifacts. The Face Synthetics dataset was specifically diversified during generation by various head poses, human identities, clothes, and such occlusions as eyeglasses and face masks.

\textbf{Annotation.} Since the face parsing annotation process is more challenging than portrait segmentation, the data is frequently marked manually or automatically with further refining. Also, annotations of existing datasets were received by inappropriate annotation rules, which prevented their usage due to our data limitations. First, almost all reviewed datasets relate beard to skin class and nostrils to nose class. Second, some contain glasses as skin and others -- annotate transparent glasses as non-transparent ones. Such factors made the skin enhancement task impossible due to further artifacts. Finally, none of them contain separate annotations for teeth.
There are other difficulties and features related to concrete ones:

\begin{itemize}
    \item LFW-PL~\cite{lfwpl} is limited to only 3 classes (background, face, hair), which unsuitable for solving our specific problems.
    \item Helen's~\cite{helen} 2,330 facial images were annotated by facial part landmarks utilizing Amazon Mechanical Truck, and then masks were generated automatically. The LaPa's~\cite{lapa} authors pointed out Helen's annotation errors.
    \item CelebAMask-HQ~\cite{celeba} ignored occlusions on its own, however, the authors \cite{faceocc} solved this problem with the dataset extension -- FaceOcc. It contains images from CelebAMask-HQ and is annotated with only one class -- occlusions (eyeglasses, tongue, makeup, and others). Regardless, FaceOcc includes a beard to the skin class. We are considered FaceOcc as CelebAMask-HQ with FaceOcc. 
    \item The iBugMask~\cite{ibugmask} contains many persons with annotated masks for only one of them. 
\end{itemize}

The mentioned datasets are inappropriate for our task due to the described and other problems such as low-resolution images, limited subject quantity, privacy concerns, and poor annotation quality. In addition to general shortcomings, other datasets lack video-conferencing domain-specific characteristics like task-specific occlusions and situations, various head poses, and domain context scenes. Inspired by the above and the necessity for an appropriate dataset for video conference apps due to its widespread use, we created a new dataset, EasyPortrait, with face parsing and portrait segmentation annotations simultaneously. We intentionally diversified the EasyPortrait by head poses, subjects, scenes, subjects' attributes such as ethnicity, and their occlusions (glasses, beards, piercing, etc.). It was annotated by all required classes for our applications, with specific rules for the skin class and occlusions especially (\cref{tabl:rules} in the
supplementary material). 

\section{EasyPortrait Dataset}
We have created a high-resolution image dataset, EasyPortrait, which contains portrait segmentation and face parsing annotations for each of 40,000 samples. This part provides our dataset creation pipeline overview, the dataset characteristics, and its splitting.

\subsection{Image Collection \& Labeling}
\label{stages}
Two crowdsourcing platforms, Yandex Toloka (for low-price labor) and ABC Elementary (for high-quality workers), were chosen for all stages of dataset creation. Our pipeline consists of two main stages: (1) the image collection stage, which is followed by validation completely realized on Yandex Toloka, and (2) the mask creation stage, which was accomplished on both platforms. At each stage, the responses of low-skilled workers were subjected to our quality control methods. A more detailed description is provided below.

\textbf{Image Collection.} 
The crowd workers' task was to take a selfie or a photo of themselves in front of the computer. As we aimed to design a diverse dataset in terms of occlusions, races, and head turns and make it suitable to solve teeth whitening problems, one of the further conditions periodically supplemented the task:
\begin{itemize}
\item Occlusions. The sent photo should contain one of such characteristics as hands in front of the face, glasses, the tongue out, headphones, or hats. 
\item Head turns. The head on the sent photo should be turned in any direction at various angles.
\item Teeth whitening. Random workers were asked to send photos with open mouths.
\item Ethnicity. We recognized the significance of having a diverse dataset of facial images and ensured participation of individuals from various countries.
\end{itemize}
Note that all workers have signed a document on the consent to the photo publication before starting the tasks and have been notified of how the data will be used (see \cref{fig: instruction} in the
supplementary material).

\textbf{Image Collection Quality Check.} We foresaw the possible dishonesty of the crowdworkers and checked all images for duplicates by image hash comparison. The image collection quality check also includes image validation, where images are reviewed for compliance with conditions such as the distinctness of the face, the head being entirely in the frame, and the clarity of the frame. Validation stage was available to crowdworkers only after training and examination tasks. Each image was checked from 3 to 5 times by different users for at least 80\% confidence.

\begin{figure*}
  \centering
  \includegraphics[width=.9\linewidth]{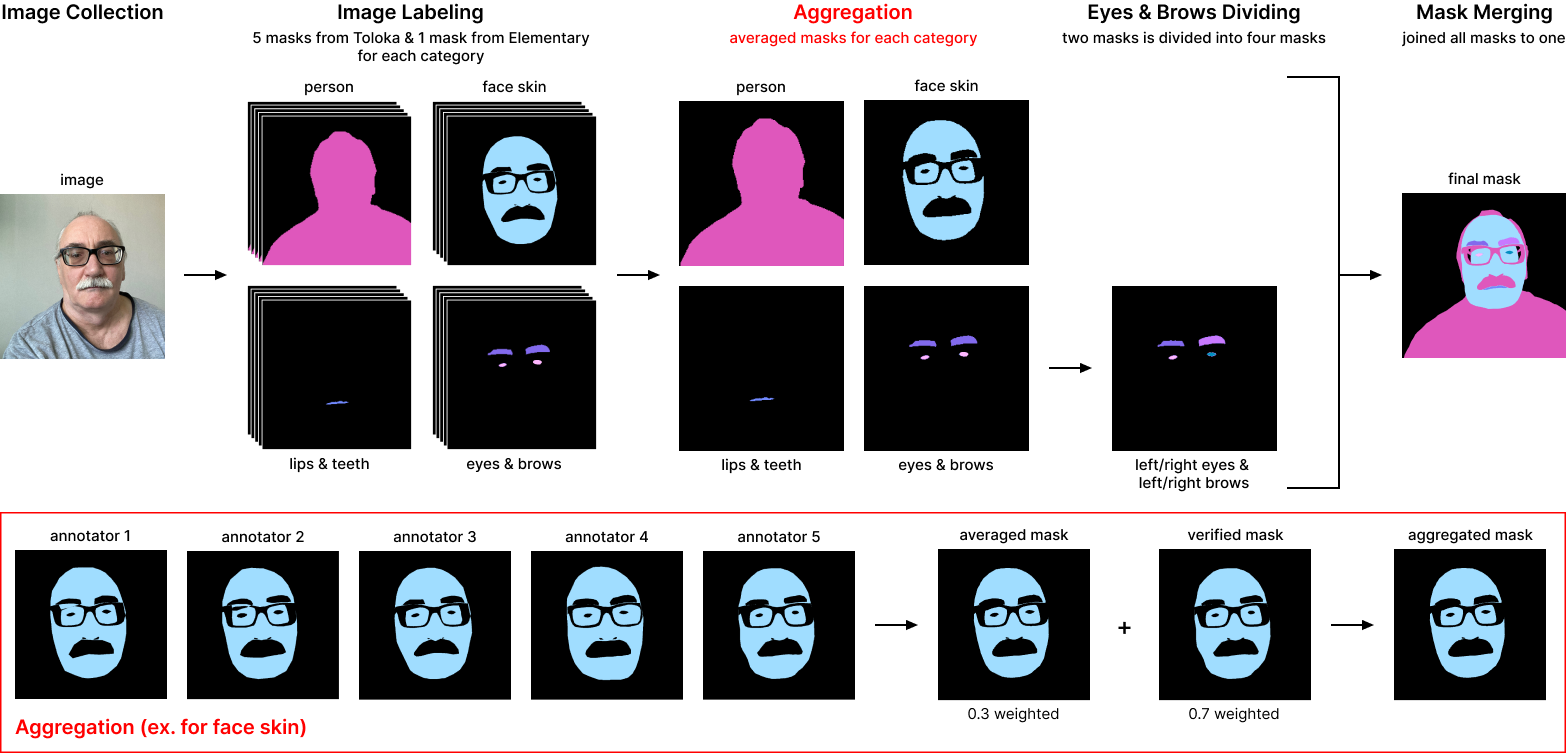}
  \caption{Example of data collection pipeline. The image was annotated with individual pair classes by 5 crowd workers. The masks are averaged with the expert-verified one and merged to obtain the final segmentation mask.}
  \label{fig: pipeline}
\end{figure*}

\textbf{Image Labeling.} The annotation of portrait segmentation usually has several ambiguous instances, such as occlusions in front of the person, hand-held items, hats, headphones, hair, and others. Face parsing masks are also unclear because of occlusions in front of the face, including tongue, hair, eyeglasses, beard, etc. The annotation rules directly affect the final segmentation masks and the model trained on them. Rules for annotating each class and processing occlusions for workers are given in \cref{tabl:rules} in the
supplementary material.

The labeling stage was divided into parts to simplify the annotation process for the workers. All images received after the collection stage were gradually sent to the annotation of individual pairs of classes: person and background, skin and occlusions (which include such things as eyeglasses, beard, tongue out, and others), eyes and brows, lips and teeth. The crowdsourcing platform's interface asks the worker to draw the polygons for the proposed instruction pair. After labeling, we separated the overall masks of eyes and brows into left and right ones using heuristics. \cref{fig: pipeline} visualizes the mask creation stages. 

\textbf{Image Labeling Quality Check.} We required all workers to complete training tasks for each pair to enhance mask quality. We analyzed crowd workers' quality in training by automatically comparing masks from expert workers under our control.

\begin{figure*}
  \centering
  \includegraphics[width=.9\linewidth]{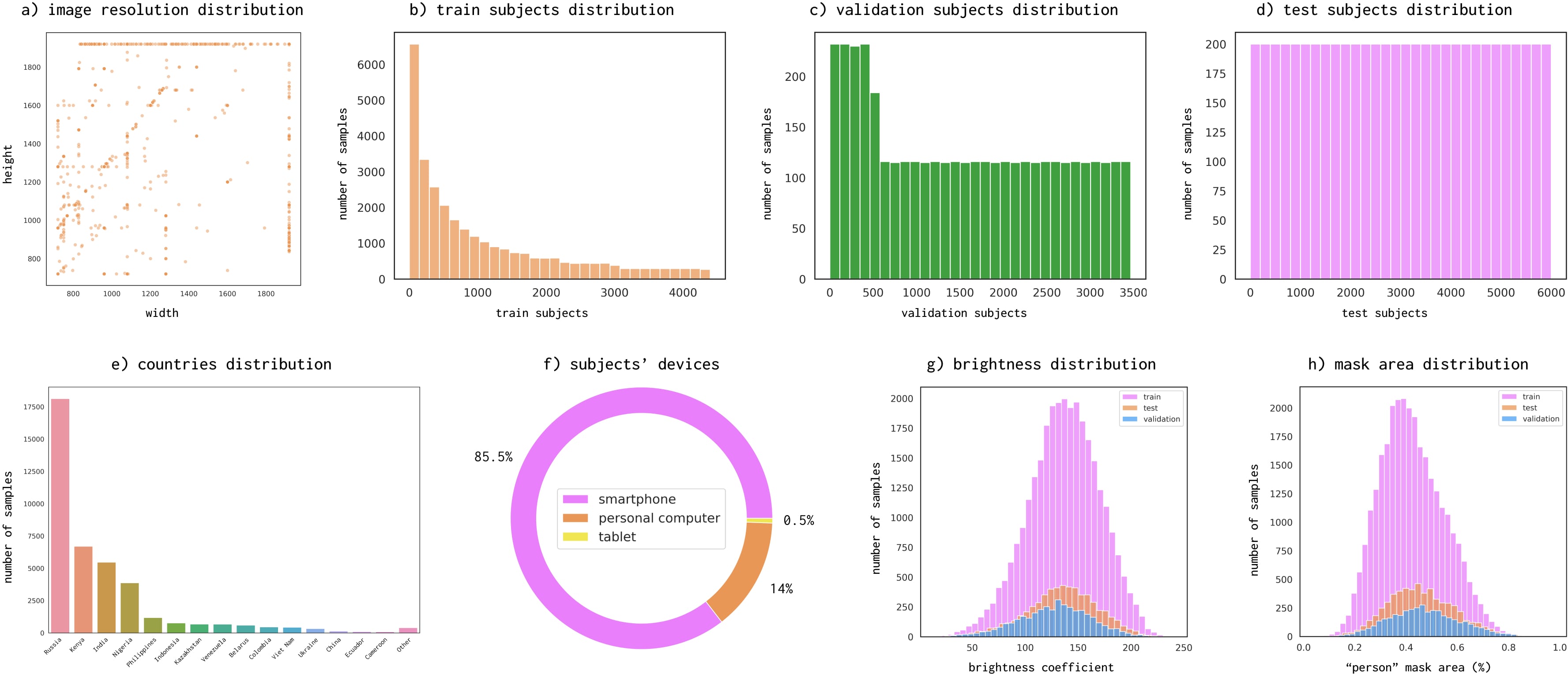}
  \caption{Image resolution, brightness, subjects, and class separability analysis. a) image resolution distribution: samples overlap with equal transparency and density reveals quantity; b), c), d) image distribution by subjects in train, validation, and test sets, respectively (each bar represents the count of images recorded by a particular subject group); e) subjects’ countries distribution; f) subjects’ devices: only smartphones, personal computers, and tablets were used while recording; g) brightness distribution; h) mask area distribution.}
  \label{fig: chars}
\end{figure*}

We requested ABC Elementary's qualified workers to label each image with the subsequent verification by the platform's experts. Due to distrust in the platform's experts, whom we did not control, and the lack of qualified annotators to provide overlap only through them, we incorporated low-skilled annotators into the pipeline with an overlap of 5. Thus, each pair of each main image was annotated by 5 crowd workers for subsequent averaging and getting the best result. Segmentation masks were created from polygons for each annotation. We aggregated all 5 markups to one segmentation mask (see bottom of the \cref{fig: pipeline}), checked IoU (Intersection over Union), and compared it to a unique threshold, selected for each pair manually\footnote{The thresholds were chosen by comparing crowd workers' masks with corresponding experts' masks from training tasks. Based on this comparison, we determined a threshold for each pair to ensure a qualitative visual result.}. Then, we averaged the received aggregated mask with verified one with weights of 0.3 and 0.7, respectively. 

\textbf{Mask Merging.} Whole masks were received by the alternate overlay of masks in the following order: person, face skin, left brow, right brow, left eye, right eye, lips, and teeth.

In addition, the decision to release the dataset to the public and ethical reasons prompted us to add the filtration stage to the end of the pipeline -- checking for children under 18, naked people, religious signs, and watermarks.

\subsection{Dataset Characteristics}
\label{characteristics}
The mean and standard deviation of images in EasyPortrait are $[0.562, 0.521, 0.497]$ and $[0.236, 0.236, 0.232]$, respectively. 

\textbf{Classes.} EasyPortrait is annotated with 9 classes, including ``background'', ``person'', ``face skin'', ``left brow'', ``right brow'', ``left eye'', ``right eye'', ``lips'', and ``teeth''. We extracted all occlusions, such as glasses, hair, hands, etc., from the skin. The beard is extracted from the skin if it is clearly defined (refer to \cref{fig: beard} in the
supplementary material for details). However, such parts of a person as headphones, car belts, and others are included in the person class to facilitate background removal and streamline the data annotation process. Nevertheless, EasyPortrait is in the process of annotation with new classes, including mouth, hair, headphones, glasses, earrings, nose, hat, neck, and beard.

\textbf{Diversity.} The proposed dataset contains images with such scenes as an office, living room, kitchen, bedroom, outdoors, car, cafe, etc. Samples in the dataset display various clothes, hats, headphones, and medical masks (see \cref{fig: samples} in the
supplementary material for examples). They are also diverse in lighting conditions, subjects' age, gender, and poses. Almost all images contain only one person, which is especially common at meetings through video conferencing services. We also collected images from regions such as Africa, Asia, India, and Europe, giving us approximate region and ethnic diversity (see \cref{fig: chars}e). Furthermore, some individuals in the pictures display emotions, such as smiling, expressing anger, sticking their tongue out, being surprise, and others.  

\textbf{Images Resolution.} Most images, namely 38,353, in EasyPortrait are FullHD: the maximum side is 1,920, and the minimum side is in the range of 835 to 1,920 (\cref{fig: chars}a). The minimal resolution is 720 × 720. \cref{fig: chars}h shows the dataset's samples' mask area distribution. 

\textbf{Dataset Quality.} We analyzed the number of points per image for each class since images were labeled by polygons. On average, each EasyPortrait image has 655 points, from which it can be concluded that the annotation is of high quality. In comparison, Helen~\cite{helen} was annotated only with 194 points per image and LaPa~\cite{lapa} with 106. \cref{fig: details} in the
supplementary material provided visual mask details. An essential part of the images was annotated with the utmost carefulness.

\subsection{Dataset Splitting}
\label{dataset-split}
All annotated images were divided into training, validation, and testing sets, with 30,000, 4,000, and 6,000 samples, respectively. Training images were received from 4,398 unique users, while validation and testing images were collected from 3,468 and 6,000 unique users, accordingly (see \cref{fig: chars}b-d). Note that the test set has the maximum amount of unique users, as we aimed to make it the most subject-diverse. It is also worth mentioning that subjects from all three sets are not intersecting, eliminating any possibility of data leakage. In addition, we added the anonymized user ID hash to the annotation file. It can be used for manual dataset splitting into train, test, and validation subsets.

\section{Ablation Study}
\label{ablation}

An ablation study was conducted to evaluate the influence of the dataset's primary characteristics. We examined the requirement for data volume and variability in head poses in our dataset by modifying these attributes, training the models, and then comparing the results with those achieved using the original dataset. In the ablation study, we utilized BiSeNetv2~\cite{bisenet}, FPN~\cite{lin2017feature}, FCN~\cite{fcn}, and Segformer-B0~\cite{segformer} for portrait segmentation and face parsing tasks. Validation and test sets remain unchanged in all ablation experiments.

\begin{figure*}
  \centering
  \includegraphics[width=\linewidth]{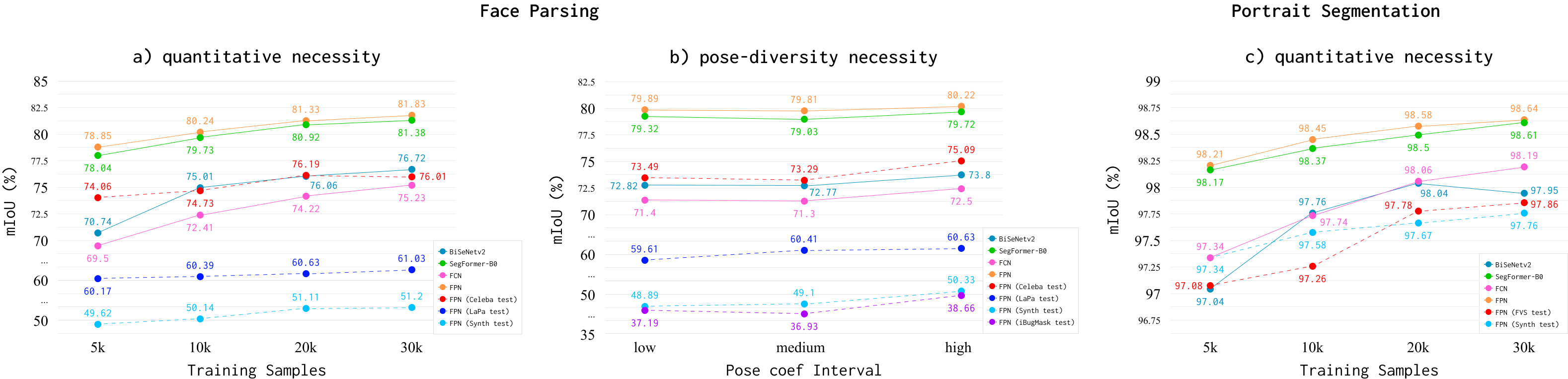}
  \caption{The impact visualization of such dataset characteristics as a) sample amount, b) head pose diversity (both for face parsing), and c) sample amount for portrait segmentation task. Solid lines correspond to models trained and tested on the EasyPortrait dataset, whereas the dotted line is the model pretrained on the EasyPortrait and tested on other datasets (see the legend for details). We evaluated all the datasets discussed in \cref{cross}; however, for datasets without significant metric changes, we did not create visualizations. Note that all the plots have different scales.}
  \label{fig: ablation}
\end{figure*}

\textbf{Quantitative Necessity.} To evaluate the impact of data quantity, we trained selected models using varying training set sizes: 30,000 (original), 20,000, 10,000, and 5,000 images. The deterministic slice was used for a train set expansion, i.e., images in the $n[i]$ set are included in the $n[i + 1]$ set. The ablation study results are provided in \cref{fig: ablation}. The quantitative necessity experiments revealed an increase in metrics as the size of the training set expanded. Both portrait segmentation and face parsing metrics show an increase with the expansion of the training set size, however the portrait segmentation improvement is less prominent than the face parsing results.

\textbf{Pose-Diversity Necessity.} We also assess the importance of the head pose by varying pose diversity in 10,000 training images. We obtained the head pose coefficients (yaw and pitch) for each image in the dataset using 3DDFA network~\cite{3ddfa_cleardusk}. For both of these coefficients, we chose three coefficient windows from homogeneous to heterogeneous pose distribution: [-7.5; 7.5], [-15; 15] [-200; 200]. Reducing head pose heterogeneity results in declining face parsing metrics (\cref{fig: ablation}). Variations in head rotations do not significantly impact portrait segmentation metrics; therefore, we did not include a plot for these experiments. 

\textbf{Cross-Dataset Ablation study.} We also conducted an additional set of experiments: we trained the FPN model on the EasyPortrait dataset with changes in data diversity and then evaluated the model on other face parsing and portrait segmentation test sets, mentioned in \cref{cross}.  

Alterations in data quantity and variations in head pose diversity have minimal impact on portrait segmentation results, while face parsing an increase in data diversity positively influences the model's metrics (\cref{fig: ablation}). On average, head pose diversity tends to have a more significant impact on results across other datasets than data quantity changes.

\section{Experiments}
\label{experiments}

The main goal of extensive base experiments is to demonstrate that the dataset has the ability to train models, achieving concurrent results without the need to simulate facial occlusion or pose variations (as in \cite{lapa} and \cite{ibugmask}, respectively). For this reason, we chose various models for our base experiments. We evaluate the models’ quality via the mean Intersection-over-Union (mIoU) metric~\cite{long2015fully}.

\begin{table}[tb]
  \caption{Evaluation results on the EasyPortrait. Column ``mIoU'' is divided into two subcolumns: face parsing and portrait segmentation tasks. For face parsing, we present mIoU metrics for each class separately, while for portrait segmentation, we provide only the overall mIoU score. We additionally trained FPN and Segformer-B0 on 224 × 224, 512 × 512, and 1024 × 1024 resolutions to demonstrate the overall increasing tendency amongst both convolutional and transformer models depending on increasing resolution.
  }
  \centering
  \scalebox{0.67}{
  \begin{tabular}{|l|c|c|c||c|c|c|c|c|c|c|c||c|}
    \hline
    \multirow{3}{*}{Model} &
      \multirow{3}{*}{Input Size} &
      \multirow{3}{*}{Model Size (MB)} &
      \multirow{3}{*}{FPS} &
      \multicolumn{9}{c|}{mIoU} \\ \cline{5-13} 
      & & & &
      \multicolumn{8}{c}{Face Parsing} &
      PS \\ \cline{5-13} 
      & & & &
      \multicolumn{1}{c}{skin} &
      \multicolumn{1}{c}{l-eye} &
      \multicolumn{1}{c}{r-eye} &
      \multicolumn{1}{c}{l-brow} &
      \multicolumn{1}{c}{r-brow} &
      \multicolumn{1}{c}{lips} &
      \multicolumn{1}{c}{teeth} &
      \multicolumn{1}{c}{overall} &
      overall \\ 
    \hline
    BiSeNetv2~\cite{bisenet} & \multirow{9}{*}{384} & 56.5 & 91.47 & 90.75 & 71.94 & 72.57 & 67.67 & 67.53 & 80.87 & 63.09 & 76.72 & 97.95 \\ 
     SegFormer-B0~\cite{segformer} & & 14.19 & 72.45 & 92.05 & 78.55 & 79.26 & 72.5 & 72.21 & 83.53 & 73.52 & 81.38 & 98.61 \\ 
     FCN + MobileNetv2~\cite{fcn} & & 31.17 & 66.07 & 90.49 & 69.95 & 70.63 & 66.29 & 66.09 & 79.23 & 59.84 & 75.23 & 98.19 \\ 
     FPN + ResNet50~\cite{lin2017feature} & & 108.91 & 58.1 & \textbf{92.28} & \textbf{79.48} & \textbf{80.08} & \textbf{72.64} & \textbf{72.47} & \textbf{84.15} & \textbf{74.09} & \textbf{81.83} & \textbf{98.64} \\ 
     DeepLabv3~\cite{chen2017rethinking} & & 260.02 & 25.65 & 91.77 & 73.78 & 74.63 & 69.61 & 69.74 & 83.42 & 70.53 & 79.11 & 98.63 \\ 
     Fast SCNN~\cite{poudel2019fastscnn} & & 6.13 & 93.89 & 88.58 & 58.42 & 58.7 & 58.68 & 58.87 & 73.16 & 44.86 & 67.56 & 97.64 \\ 
      DANet~\cite{fu2019dual} & & 190.29 & 42.43 & 91.8 & 74.01 & 74.93 & 70.01 & 69.75 & 83.7 & 70.8 & 79.3 & 98.63 \\ 
     EHANet~\cite{luo2020ehanet} & & 44.81 & 132.78 & 89.68 & 68.87 & 69.26 & 63.6 & 63.82 & 73.98 & 52.05 & 72.56 & - \\ 
     SINet~\cite{park2019sinet} & & 0.13 & 134.18 & - & - & - & - & - & - & - & - & 93.32 \\ 
     ExtremeC3Net~\cite{park2019extremec3net} & & 0.15 & 71.75 & - & - & - & - & - & - & - & - & 96.54 \\ 
     \hline \hline
     SegFormer-B0 & \multirow{2}{*}{224} & 14.9 & 74.84 & 90.19 & 68.59 & 70.46 & 65.79 & 65.72 & 77.94 & 60.66 & 74.83 & 98.17 \\ 
     \cline{1-1} \cline{3-13} 
     FPN + ResNet50 & & 108.91 & 61.56 & 90.6 & 69.67 & 71.88 & 65.84 & 65.64 & 78.94 & 62.95 & \textbf{75.6} & \textbf{98.31} \\ 
     \cline{1-1} \cline{3-13} 
     \hline
     SegFormer-B0 & \multirow{2}{*}{512} & 14.9 & 65.88 & 92.5 & 81.03 & 81.18 & 74.31 & 74.08 & 84.87 & 78.14 & 83.19 & \textbf{98.66} \\ 
     \cline{1-1} \cline{3-13} 
     FPN + ResNet50 & & 108.91 & 53.14 & 92.55 & 81.55 & 81.47 & 74.33 & 74.38 & 85.27 & 77.77 & \textbf{83.33} & 98.64 \\ 
     \cline{1-1} \cline{3-13} 
     \hline
     SegFormer-B0 & \multirow{2}{*}{1024} & 14.9 & 62.9 & 93.13 & 84.2 & 83.97 & 76.41 & 76.12 & 86.88 & 83.2 & \textbf{85.42} & \textbf{98.74} \\ 
     \cline{1-1} \cline{3-13} 
     FPN + ResNet50 & & 108.91 & 52.34 & 92.94 & 84.55 & 84.24 & 76.11 & 76.11 & 86.93 & 82.62 & 85.37 & 98.54 \\ 
     \hline
  \end{tabular}}
\label{tabl:exps}
\end{table}

\subsection{Base Experiments}
\label{base}
\textbf{Separation on Two Tracks.} We split our experiments into two tracks -- portrait segmentation and face parsing -- to transparently compare EasyPortrait with other datasets separately. This division is also necessary to avoid ambiguity and ensure the obtained metrics are representative of both tasks. The portrait segmentation is based on two EasyPortrait classes (``background'' and ``person''), whereas the face parsing masks include eight classes (``background'', ``skin'', ``left brow'', ``right brow'', ``left eye'', ``right eye'', ``lips'' and ``teeth''). For portrait segmentation, we defined all classes of EasyPortrait except the background as a person, while for face parsing, we designated the person class as the background. The model configuration and training process are identical for both tasks, except for the number of classes in the decoder model's head.

\textbf{Models.} We prioritized lightweight architectures for easy integration into videoconferencing apps, enabling real-time use. As general segmentation architectures, we selected BiSeNetv2~\cite{bisenet}, DeepLabv3~\cite{chen2017rethinking}, FPN~\cite{lin2017feature}, FCN~\cite{fcn}, DANet~\cite{fu2019dual}, and Fast SCNN~\cite{poudel2019fastscnn} models. We utilized Segformer-B0~\cite{segformer} to assess the performance of the transformer model on the proposed dataset. Besides the aforementioned widespread segmentation architectures, we also experimented with models specifically designed for portrait segmentation and face parsing. For this purpose, we chose the SINet~\cite{park2019sinet} and ExtremeC3Net~\cite{park2019extremec3net} for the first one and EHANet~\cite{luo2020ehanet} model for the second. 

We trained each of these networks for 100 epochs with batch size 32. AdamW~\cite{adamw} was used as an optimizer and learning rate with the initial value of 0.0002. The learning rate changes according to the polynomial learning rate scheduler with factor 1.0 by default.

\textbf{Augmentations and Images Resolution.} Images and segmentation masks were resized to the maximum side of 384 with aspect ratio preservation and symmetrically padded to square. We used bilinear interpolation for image resizing, while nearest neighbor interpolation was applied to masks to maintain consistency among classes. At last, photometric distortion was used with a brightness delta of 16, a contrast in the range [0.5, 1.0], saturation in the range [0.5, 1.0], and a hue delta of 5. 

The results of our experiments are presented in the \cref{tabl:exps}. All the models trained on our dataset achieve high metrics, with the FPN model outperforming others in face parsing and portrait segmentation tasks.

\subsection{Cross-Dataset Evaluation}
\label{cross}
We conduct cross-dataset evaluation to compare our dataset with existing ones in 2 domains -- face parsing and portrait segmentation.

\textbf{Experiments Configuration.} We train the FPN model for two segmentation tasks on each dataset. All datasets' samples were exposed to resizing to fixed 384 × 384 shape and base augmentations pipeline described in~\cref{base}. The training process and model configuration are the same as the base experiments for both tasks.

\textbf{Portrait Segmentation.} Besides our dataset, the model was trained and tested on HumanSeg14K~\cite{paddleseg}, Face Synthetics~\cite{synthetic}, and FVS~\cite{ConferenceVideoSegmentationDataset} portrait segmentation datasets. We couldn't include the EG-1800~\cite{eg1800} and the AiSeg~\cite{aiseg} datasets due to a lack of images on the public shared sources and inappropriate samples, respectively.

Some preprocessing steps were applied to each of the datasets:
\begin{itemize}
    \item We led the EasyPortrait's class ``person'' to a consistent appearance by labeling others classes (without ``background'') as ``person'' class.
    \item FVS~\cite{ConferenceVideoSegmentationDataset} is announced as a segmentation dataset; however, the provided masks are not binary, so we binarize them. We found out that pixel values are mostly scattered near 0 or 255; therefore, the average threshold of 127 was chosen to separate the ``person'' and ``background'' classes. The original dataset has been split into 1,326 training and 935 testing samples. We randomly sampled 200 images from the training set to design the validation.
    \item HumanSeg14K~\cite{paddleseg} dataset was divided into the training, validation, and test parts with 8,770, 2,431, and 2,482 samples, respectively. 
    \item Similar to EasyPortrait's preprocessing, we prepared the Face Synthetics~\cite{paddleseg} dataset to portrait segmentation masks. We randomly picked 75,000 training, 15,000 testing, and 10,000 validation samples. 
\end{itemize}

\textbf{Face Parsing.} As far as the EasyPortrait skin class was annotated by the unique rules and most face parsing datasets are not annotated with the teeth class, we selected only 6 classes for cross-data evaluation: ``background'', ``left brow'', ``right brow'', ``left eye'', ``right eye'', ``lips''. We adopted the original annotations of face parsing datasets to the target ones:
\begin{itemize}
    \item Such EasyPortrait classes as ``teeth'', ``person'' and ``skin'' classes were mapped to the ``background''.
    \item Since lips of the CelebAMask-HQ dataset are divided into two classes: ``lower lip'' and ``upper lip'', we combined them into one ``lips'' class. The remaining classes are considered as background. All datasets below were preprocessed in the same way as CelebAMask-HQ~\cite{celeba}. We divided the CelebAMask-HQ dataset into 22,500 training, 3,000 validation, and 4,500 test samples. 
    \item The LaPa~\cite{lapa} dataset was originally split into 18,167 training, 2,000 validation, and 2,000 test samples.
    \item Originally, images from iBugMask~\cite{ibugmask} were split into 21,866 training and 1,000 testing examples. The validation set was randomly sampled from the training set and contains 1,866 images. Note that iBugMask~\cite{ibugmask} contains images with a bounding box for the face in the provided mask. To avoid parsing other faces, we crop them as described in the original paper.
    \item The Face Synthetics dataset was distributed into 75,000 training, 10,000 validation, and 15,000 test samples by us. 
\end{itemize}

\begin{table}[tb]
  \caption{Cross-dataset evaluation results. Each cell value contains mean (among classes) IoU metrics for the corresponding training and testing sets pair. Train (test) average mIoU represents the overall mean IoU value on the listed testing (training) sets. There is a direct relationship between the high train average mIoU metric and the dataset's generalization ability to other distributions. The low test average mIoU metric value reflects the dataset's complexity, as a model pre-trained on a different training set struggles to achieve a high metric. This metric indicates the dataset's suitability as a benchmark for face parsing or portrait segmentation tasks. We highlighted the best metric in each column to emphasize the dataset's ability to generalize to other distributions. The best metric in all columns except the last one was chosen, excluding diagonal values. (upper table) The FPN model, trained on the EasyPortrait dataset for portrait segmentation, achieves state-of-the-art results and surpasses FVS results even on their own test set. (lower table) Despite the limited videoconferencing domain, we achieved concurrent results in the face parsing task.
  }
  \centering
  \scalebox{0.798}{
  \begin{tabular}{|lcccccc|}
    \hline
    \multicolumn{7}{|c|}{Portrait Segmentation} \\ 
    \hline
    \multicolumn{1}{|l}{} & \multicolumn{1}{|l|}{} & \multicolumn{4}{c|}{Tested} & \multicolumn{1}{c|}{} \\ 
    \hline
    \multicolumn{1}{|l}{} & \multicolumn{1}{|l|}{Dataset} & \multicolumn{1}{c|}{EasyPortrait (ours)} & \multicolumn{1}{c|}{FVS} & \multicolumn{1}{c|}{HumanSeg14K} & \multicolumn{1}{c|}{Face Synthetics} & \multicolumn{1}{c|}{Train avg. mIoU} \\ 
    \hline
    \multicolumn{1}{|l|}{\multirow{4}{*}{Trained}} & \multicolumn{1}{l|}{EasyPortrait (ours)} & \multicolumn{1}{c|}{\cellcolor{LightCyan}98.64} & \multicolumn{1}{c|}{\textbf{97.86}} & \multicolumn{1}{c|}{\textbf{93.18}} & \multicolumn{1}{c|}{\textbf{97.76}} & \multicolumn{1}{c|}{\textbf{96.86}}  \\ 
    \hhline{~|-|-|-|-|-|-}
    \multicolumn{1}{|l|}{} & \multicolumn{1}{l|}{FVS~\cite{ConferenceVideoSegmentationDataset}} & \multicolumn{1}{c|}{79.05} & \multicolumn{1}{c|}{\cellcolor{LightCyan}96.24} & \multicolumn{1}{c|}{90.6} & \multicolumn{1}{c|}{80.36} & \multicolumn{1}{c|}{86.56}  \\ 
    \hhline{~|-|-|-|-|-|-}
    \multicolumn{1}{|l|}{} & \multicolumn{1}{l|}{HumanSeg14K~\cite{paddleseg}} & \multicolumn{1}{c|}{76.01} & \multicolumn{1}{c|}{96.23} & \multicolumn{1}{c|}{\cellcolor{LightCyan}97.53} & \multicolumn{1}{c|}{71.66} & \multicolumn{1}{c|}{85.35}  \\ 
    \hhline{~|-|-|-|-|-|-}
    \multicolumn{1}{|l|}{} & \multicolumn{1}{l|}{Face Synthetics~\cite{synthetic}} & \multicolumn{1}{c|}{\textbf{84.99}} & \multicolumn{1}{c|}{57.14} & \multicolumn{1}{c|}{57.87} & \multicolumn{1}{c|}{\cellcolor{LightCyan}99.44} & \multicolumn{1}{c|}{74.86}  \\ 
    \hline
    \multicolumn{1}{|l|}{} & \multicolumn{1}{l|}{Test avg. mIoU} & \multicolumn{1}{c|}{84.67} & \multicolumn{1}{c|}{86.87} & \multicolumn{1}{c|}{84.8} & \multicolumn{1}{c|}{87.31} & \multicolumn{1}{c|}{} \\ 
  \hline
  \end{tabular}}
\scalebox{0.71}{
  \begin{tabular}{llcccccc}
    \hline
    \multicolumn{8}{|c|}{Face Parsing} \\ \hline
    \multicolumn{1}{|l}{} & \multicolumn{1}{|l|}{} & \multicolumn{5}{c|}{Tested} & \multicolumn{1}{c|}{} \\ \hline
    \multicolumn{1}{|l|}{} & \multicolumn{1}{l|}{Dataset} & \multicolumn{1}{c|}{EasyPortrait (ours)} & \multicolumn{1}{c|}{CelebAMask-HQ} & \multicolumn{1}{c|}{iBugMask} & \multicolumn{1}{c|}{Face Synthetics} & \multicolumn{1}{c|}{LaPa} & \multicolumn{1}{c|}{Train avg. mIoU} \\ \hline
    \multicolumn{1}{|l|}{\multirow{5}{*}{Trained}} & \multicolumn{1}{l|}{EasyPortrait (ours)} & \multicolumn{1}{c|}{\cellcolor{LightCyan}81.51} & \multicolumn{1}{c|}{76.01} & \multicolumn{1}{c|}{39.0} & \multicolumn{1}{c|}{\textbf{51.2}} & \multicolumn{1}{c|}{61.03} & \multicolumn{1}{c|}{61.75} \\
    \hhline{~|-|-|-|-|-|-|-}
    \multicolumn{1}{|l|}{} & \multicolumn{1}{l|}{CelebAMask-HQ~\cite{celeba}} & \multicolumn{1}{c|}{66.17} & \multicolumn{1}{c|}{\cellcolor{LightCyan}83.41} & \multicolumn{1}{c|}{\textbf{54.74}} & \multicolumn{1}{c|}{46.6} & \multicolumn{1}{c|}{60.62} & \multicolumn{1}{c|}{62.31}  \\ 
    \hhline{~|-|-|-|-|-|-|-}
    \multicolumn{1}{|l|}{} & \multicolumn{1}{l|}{iBugMask~\cite{ibugmask}} & \multicolumn{1}{c|}{61.58} & \multicolumn{1}{c|}{\textbf{79.1}} & \multicolumn{1}{c|}{\cellcolor{LightCyan}64.59} & \multicolumn{1}{c|}{44.42} & \multicolumn{1}{c|}{\textbf{66.3}} & \multicolumn{1}{c|}{63.19} \\
    \hhline{~|-|-|-|-|-|-|-}
    \multicolumn{1}{|l|}{} & \multicolumn{1}{l|}{Face Synthetics~\cite{synthetic}} & \multicolumn{1}{c|}{55.55} & \multicolumn{1}{c|}{40.67} & \multicolumn{1}{c|}{18.84} & \multicolumn{1}{c|}{\cellcolor{LightCyan}83.12} & \multicolumn{1}{c|}{42.63} & \multicolumn{1}{c|}{48.16}   \\ 
    \hhline{~|-|-|-|-|-|-|-}
    \multicolumn{1}{|l|}{} & \multicolumn{1}{l|}{LaPa~\cite{lapa}} & \multicolumn{1}{c|}{\textbf{68.56}} & \multicolumn{1}{c|}{73.92} & \multicolumn{1}{c|}{47.66} & \multicolumn{1}{c|}{48.05} & \multicolumn{1}{c|}{\cellcolor{LightCyan}79.02} & \multicolumn{1}{c|}{\textbf{63.44}}  \\ 
    \hline
    \multicolumn{1}{|l|}{} & \multicolumn{1}{l|}{Test avg. mIoU} & \multicolumn{1}{c|}{66.67} & \multicolumn{1}{c|}{70.62} & \multicolumn{1}{c|}{44.97} & \multicolumn{1}{c|}{54.68} & \multicolumn{1}{c|}{61.92} & \multicolumn{1}{c|}{}  \\ 
  \hline
  \end{tabular}}
\label{tabl:cross_val}
\end{table}

\textbf{Results.} The cross-dataset evaluation results in \cref{tabl:cross_val} demonstrate that the EasyPortrait has the best generalization capability regarding mIoU metrics on each portrait segmentation test set. Due to the reduced list of classes, the quantitative assessment provides limited insights into the dataset’s applicability for the face parsing task. Besides, our domain is slightly limited\footnote{Images in other datasets are more heterogeneous in context, displaying multiple people and different activities. EasyPortrait consistently shows a single person in front of a computer or phone.}, which prevented achieving more optimistic results.

\section{Conclusion}
In this paper, we proposed a large-scale image dataset for portrait segmentation and face parsing tasks. It consists of 40,000 photos of ordinary people predominantly indoors, and each image is provided with a 9-classes high-quality semantic mask. Our dataset can be used in several beautification and segmentation tasks, such as background removal, face skin enhancement, and teeth whitening, which can improve user experience in video conferencing apps. We provide extensive experiments on different models and cross-dataset comparisons for both described tasks. We also conducted an ablation study revealing the importance of such dataset characteristics as data quantity and head pose diversity for training a precise and robust model. For future work, we plan to add several occlusions to the annotation and improve the dataset by including additional classes, such as mouth, hair, headphones, glasses, earrings, nose, hat, neck, and beard. 

%
%
\bibliographystyle{splncs04}
\bibliography{eccv/eccv}
\clearpage
\setcounter{page}{1}
\appendix


\setcounter{figure}{4}
\setcounter{table}{3}

\makeatletter







\appendix
\section{Supplementary material}
\label{sec:supplementary}

\begin{figure*}[htp]
  \centering
  \includegraphics[width=\linewidth]{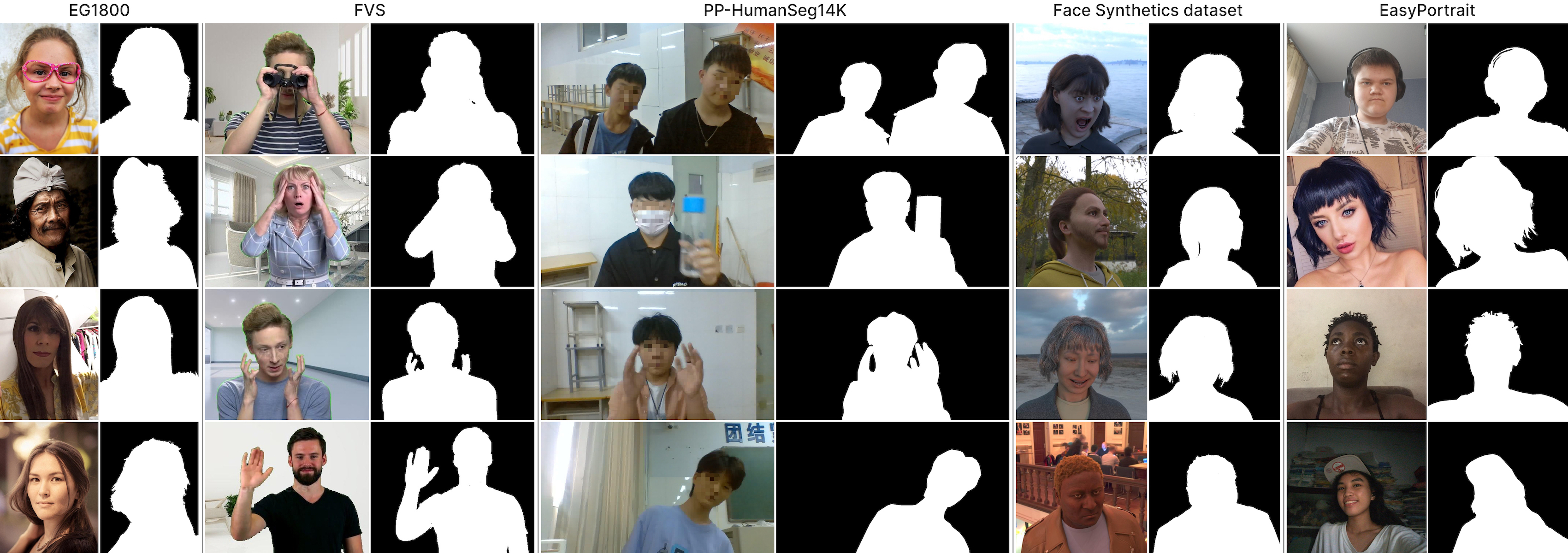}
  \caption{Visual comparison of existing portrait segmentation datasets. One can notice high-frequency details (e.g. hair) in segmentation masks in samples from our dataset. The AiSeg~\cite{aiseg} dataset is not considered since it provides the extracted foreground images without corresponding annotation mask.}
  \label{fig: ps}
\end{figure*}

\begin{figure*}[htp]
  \centering
  \includegraphics[width=\linewidth]{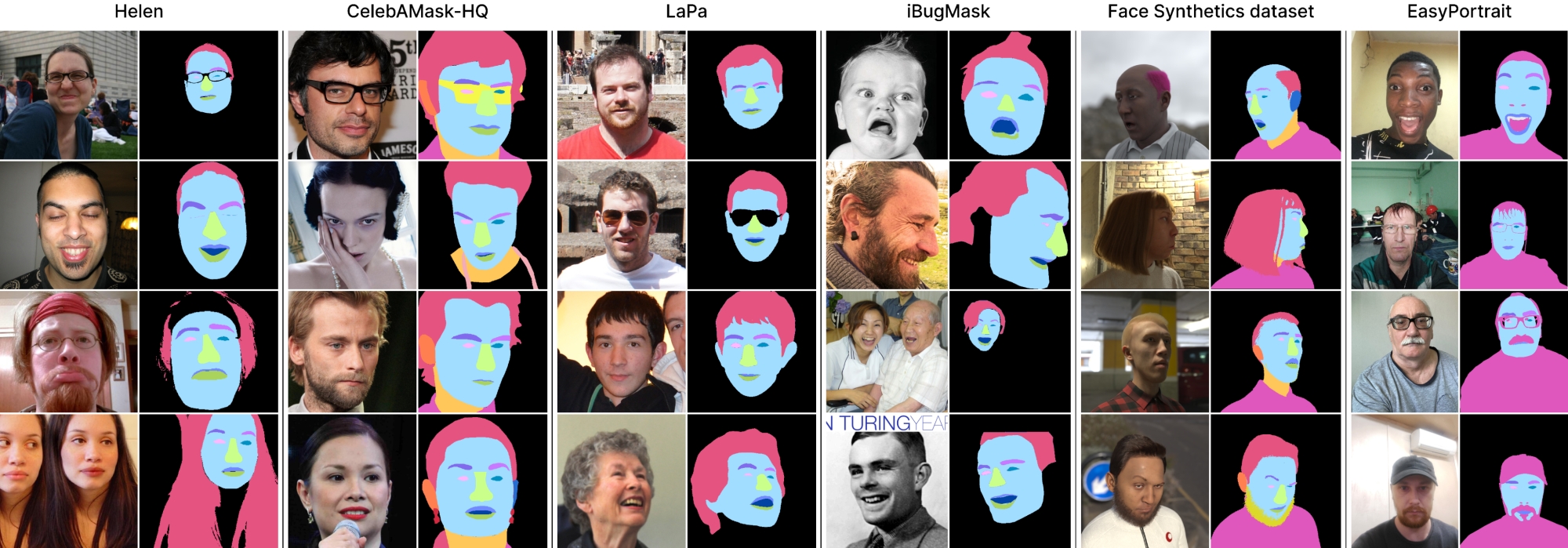}
  \caption{Visual comparison of existing face parsing datasets. Only Face Synthetics~\cite{synthetic} and EasyPortrait datasets can be used to solve background removal and face enhancement problems. None of them except EasyPortrait can be used in the teeth whitening task. We don't include LFW-PL~\cite{lfwpl} and FaceOcc~\cite{faceocc} datasets in the visualization due to the lack of classes and the need for preprocessing, respectively.}
  \label{fig: fp}
\end{figure*}

\begin{figure*}
  \centering
  \includegraphics[width=.8\linewidth]{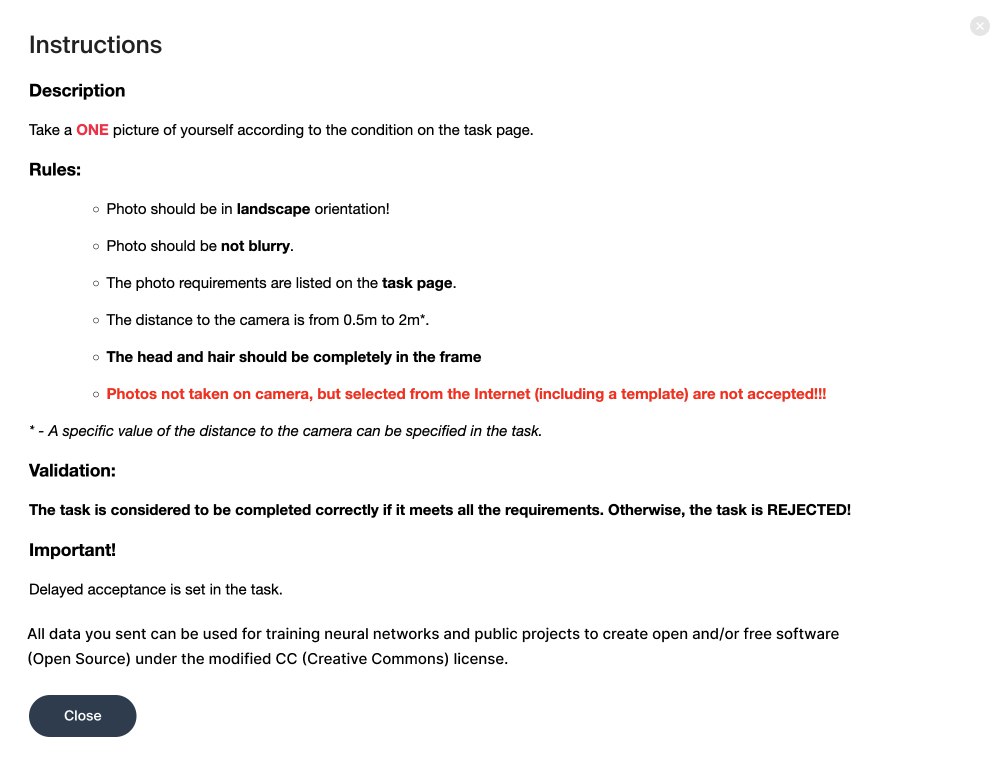}
  \caption{Example of instruction for the task with notification of further use of data for training neural networks.}
  \label{fig: instruction}
\end{figure*}

\begin{figure*}
  \centering
  \includegraphics[width=.9\linewidth]{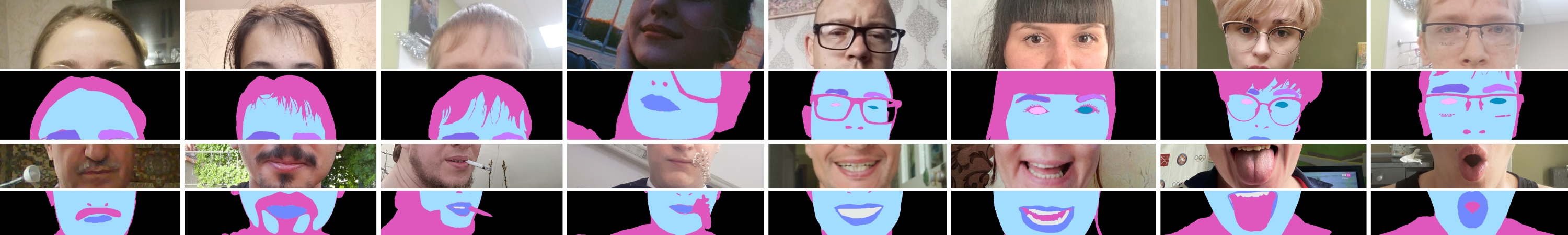}
  \caption{Examples of facial details in the EasyPortrait dataset. Samples contain many different facial expressions, with open and closed mouth, with or without teeth shown. Note that obstacles in front of the face (like glasses and a cigarette) are annotated as a class ``person".}
  \label{fig: details}
\end{figure*}

\begin{figure*}[htp]
  \centering
  \includegraphics[width=\linewidth]{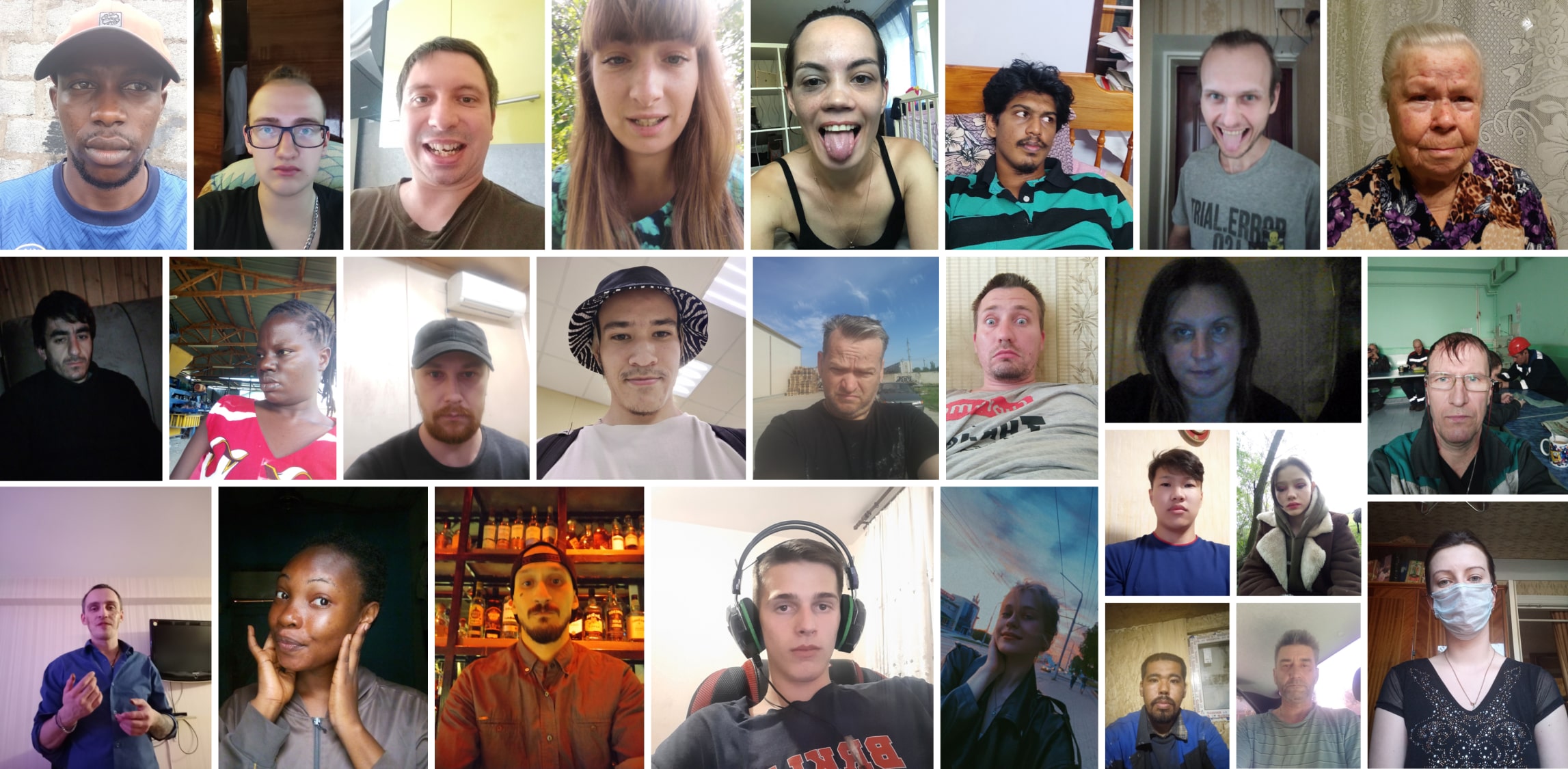}
  \caption{Samples from EasyPortrait dataset. Note that images have different orientation and height-width ratio. Some of the pictures are overexposed, while other are underexposed.}
  \label{fig: samples}
\end{figure*}

\begin{figure*}[htp]
  \centering
  \includegraphics[width=\linewidth]{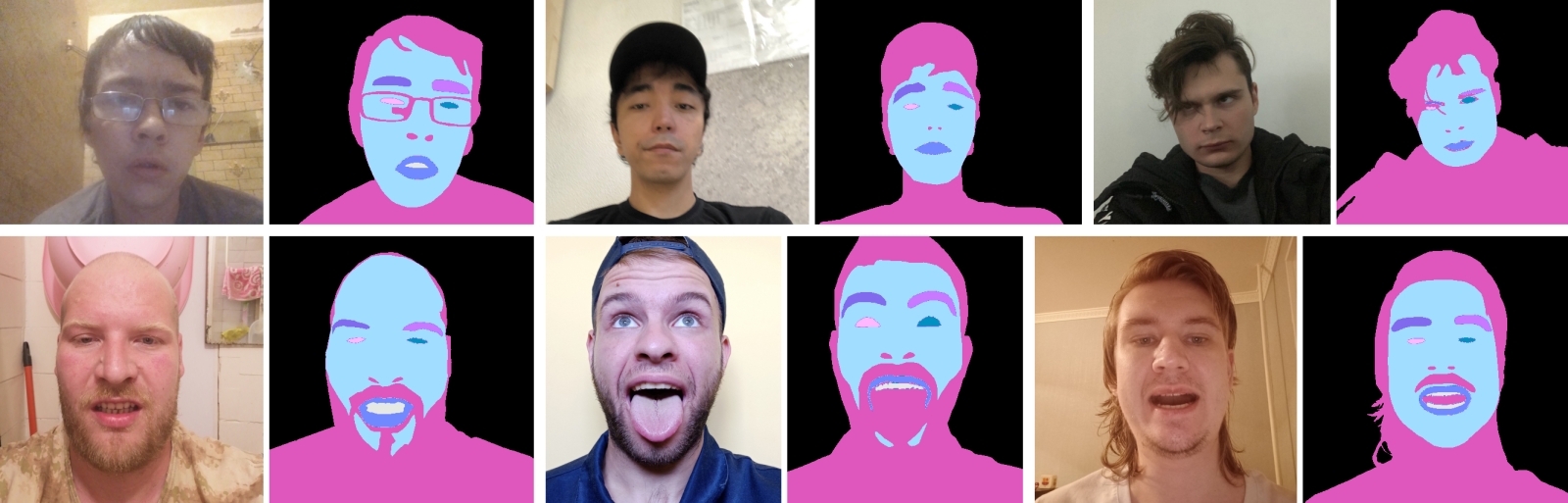}
  \caption{Visualization of the beard annotation rules. (up) The beard is included in the skin if it is a separate hair or barely noticeable. (bottom) The beard is excluded from the skin if it is clear.}
  \label{fig: beard}
\end{figure*}

\begin{figure*}[htp]
  \centering
  \includegraphics[width=\linewidth]{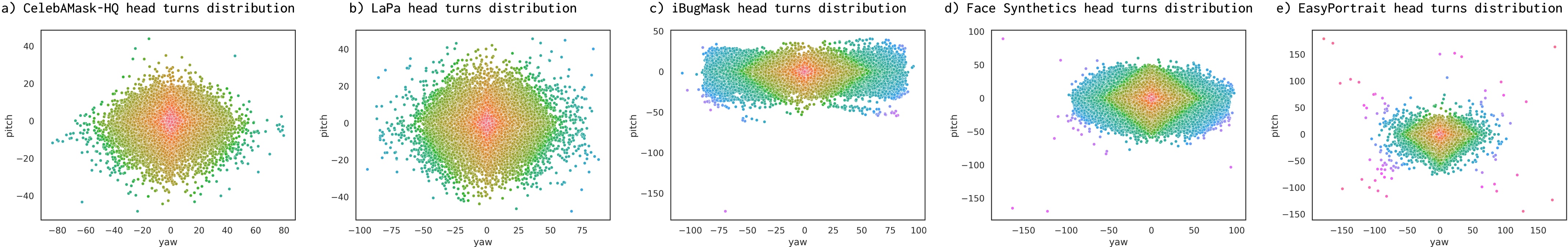}
  \caption{Head turns distributions for several face parsing and portrait segmentation datasets, including EasyPortrait. Yaw and pitch coefficients were obtained via 3DDFA network~\cite{3ddfa_cleardusk}.}
  \label{fig: turns}
\end{figure*}

\begin{table*}
\centering
\caption{EasyPortrait annotators rules. Visual examples are provided in \cref{fig: details}.}
\scalebox{0.73}{
\begin{tabular}{|p{0.75in}|p{5.75in}|}
\hline
Class & Rules\\
\hline\hline
Person & -- headphones and things in front of the person are defined as a person's class \newline -- individual hairs and all empty areas closed by a person are not included in the person class\\
\hline
Eyebrows & stand out along a strict border, excluding individual hairs\\
\hline
Eyes & distinguished by whites, excluding eyelids and eyelashes\\
\hline
Skin & -- the skin class should affect only skin without hair, eyes, and other face attributes \newline -- the boundaries of the skin of the face or person should be highlighted logically on overexposed or darkened photos \newline -- the rare bristle also considered skin \newline -- ears, second chin, and nostrils are not included in the skin class\\
\hline
Teeth & teeth and everything else in the open mouth stand out separately, the latter as an occlusion\\
\hline
Occlusions & -- makeups and piercing are defined as occlusions \newline -- the part of eyeglasses, which cover skin, should be annotated as occlusion, including sunglasses and glare on clear glasses \newline -- beard with a strict border are considered occlusion \newline -- the tongue out of the mouth should be annotated as occlusion\\
\hline
\end{tabular}}
\label{tabl:rules}
\end{table*}
\end{document}